%% file: Template.tex
\documentclass{article}
\usepackage{spconf,amsmath,graphicx}
\usepackage{amssymb}
\usepackage{booktabs}
\usepackage{dsfont}
\usepackage{hyperref}

\usepackage{xcolor}

\title{DUAL-SPACE AUGMENTED INTRINSIC-LORA FOR WIND TURBINE SEGMENTATION}
\name{Shubh Singhal$^{1,2,*}$, Raül~Pérez-Gonzalo$^{1,3,*}$, Andreas~Espersen$^{3}$ and Antonio~Agudo$^{1}$\thanks{This work has been supported by the Innovation Fund Denmark under 2021 ID1044-0044A, by the project GRAVATAR PID2023-151184OB-I00 funded by MCIU/AEI/10.13039/501100011033 and by ERDF, UE; and by the project GreenVAR of the Fundación Ramón Areces. 
$^{*}$Equal contribution.}}
\address{$^{1}$Institut de Robòtica i Informàtica Industrial, CSIC-UPC, Barcelona, Spain\\$^{2}$Université de Toulon, La Garde, France\\$^{3}$Wind Power LAB, Copenhagen, Denmark}
\begin{document}
%
\maketitle
\begin{abstract}
Accurate segmentation of wind turbine blade (WTB) images is critical for effective assessments, as it directly influences the performance of automated damage detection systems. Despite advancements in large universal vision models, these models often underperform in domain-specific tasks like WTB segmentation. To address this, we extend Intrinsic LoRA for image segmentation, and propose a novel dual-space augmentation strategy that integrates both image-level and latent-space augmentations. The image-space augmentation is achieved through linear interpolation between image pairs, while the latent-space augmentation is accomplished by introducing a noise-based latent probabilistic model. Our approach significantly boosts segmentation accuracy, surpassing current state-of-the-art methods in WTB image segmentation.
\end{abstract}
\begin{keywords}
Latent-space Augmentation, Diffusion Models, LoRA, Image Segmentation, Wind Turbine Blade
\end{keywords}

\input{Chapters/Introduction}
\input{Chapters/Methods}

\input{Chapters/ExperimentalResults}
\input{Chapters/Conclusions}

\bibliographystyle{IEEEbib}
\bibliography{strings}

\end{document}

%% file: Chapters/Introduction.tex
\vspace{-0.1cm}
\section{Introduction}
\label{sec:intro}
\vspace{-0.15cm}

Operational damages to wind turbine blades (WTBs) can greatly impact their efficiency~\cite{delamination-efficiency} and may even lead to complete failure~\cite{turbine-failure}. Regular visual inspections and preventive maintenance are crucial to ensure timely repairs. These inspections are typically conducted using drones that capture high-resolution images, allowing for detailed analysis to guide maintenance decisions~\cite{HPL}. As the wind energy sector rapidly expands, the need for automated WTB assessment solutions is increasing, with image segmentation emerging as a key image processing task in this process~\cite{bunet}.


Deep learning methods, particularly convolutional neural networks (CNNs), have driven significant advances in image segmentation research. Encoder-decoder architectures~\cite{deeplab,sw} have become foundational frameworks by effectively capturing and reconstructing spatial relationships. Some notable models like
DeepLabv3+~\cite{deeplabv3+} and ResNeSt~\cite{resnest} achieve remarkable success by utilizing atrous convolutions and multi-scale feature extraction techniques. The integration of attention mechanisms with CNNs, such as U-NetFormer~\cite{unetformer}, have further enhanced segmentation capabilities by improving global context understanding~\cite{attention2,bmvc}. In the realm of WTB segmentation, these advancements have inspired tailored models like BU-Net~\cite{bunet}, which incorporates a post-processing hole-filling algorithm to refine segmentation results.
    
There has been growing interest in universal image segmentation models trained in a zero-shot manner using vast amounts of data, particularly large vision models like SAM \cite{sam} and DINO \cite{dino}. These models employ self-supervised vision transformers that learn meaningful representations from data through self-attention mechanisms. However, in practical applications, these universal segmentation methods often underperform compared to state-of-the-art models and cannot be trained directly due to the lack of comprehensive datasets. Intrinsic LoRA~\cite{intrinsic-lora} presents a promising approach by fine-tuning generative models trained on large datasets, enabling supervised learning with minimal labeled data.

\begin{figure}[t!]
  \centering
  \includegraphics[width=\linewidth]{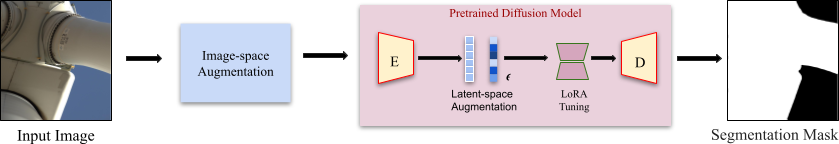}
  \vspace{-0.8cm}
\caption{\textbf{General schema of Segmentation-based Intrinsic LoRA (SI-LoRA) with dual-space augmentation.} }
\label{fig:basic_pipeline}
\vspace{-0.6cm}
\end{figure}

In this work, we extend Intrinsic LoRA for image segmentation and demonstrate its effectiveness in a real-world application. Specifically, we adapt pretrained Stable Diffusion models~\cite{stable-diffusion} for WTB segmentation by modifying the segmentation masks to meet the dimensional requirements of the pretrained model. These initial results, however, produce suboptimal performance. To address this, we explore several augmentation techniques. Initially, we apply traditional data augmentation methods to the input images~\cite{mixup,cutmix,zhang2021objectaug}, which prove particularly effective. Then, inspired by prior research that stabilizes the training of generative models in the latent space~\cite{Sønderby2016a,jenni2019stabilizing,dual-space-augmentation}, we introduce a Bayesian adaptation of Intrinsic LoRA for image segmentation, modeling the latent vectors in a probabilistic augmented framework. By integrating both image-level and latent-space augmentations (see Fig.~\ref{fig:basic_pipeline}), our dual-space augmentation approach substantially improves segmentation performance, surpassing state-of-the-art methods in WTB segmentation by a large margin.

%% file: Chapters/Methods.tex

\section{Methodology}

This section outlines our adaptation of the Intrinsic LoRA~\cite{intrinsic-lora} method for image segmentation. We begin by reviewing the core principles of Intrinsic LoRA and its application in extracting image intrinsics. Next, we proceed by illustrating how we tailored this approach to create segmentation maps. Finally, we propose a novel dual-space augmentation method that operates in both image and latent spaces.
\vspace{-0.35cm}

\subsection{Intrinsic LoRA} \label{sec:i-lora} \vspace{-0.15cm}

Intrinsic-LoRA~\cite{intrinsic-lora} harnesses the implicit understanding of image intrinsics within generative models to produce high-quality supervised outputs. By introducing learnable LoRA~\cite{lora} adaptors $\boldsymbol{\theta}$, an image-to-image generative diffusion model can be fine-tuned with minimal labeled samples to generate the desired outputs $\mathbf{y} \in \mathds{R}^{H \times W \times 3}$.

Given a pretrained Stable Diffusion model~\cite{stable-diffusion}, the input image $\mathbf{x} \in \mathds{R}^{H \times W \times 3}$ is encoded by the encoder $E$ to a lower-dimensional latent space $\mathbf{z}^{(E)}_{\mathbf{x}} = E(\mathbf{x})$. The obtained latent vector $\mathbf{z}^{(E)}_{\mathbf{x}}$ is fed to the denoising U-Net~\cite{unet} model $U_{\boldsymbol{\theta}} $ along with a text prompt $t$. This prompt is the image intrinsic to be extracted like "depth", "normal" and so forth, and is encoded by a pretrained CLIP~\cite{clip} tokenizer $T$, obtaining the transformed output latent vector $ \mathbf{z}^{(U)}_{\mathbf{x}}  = U_{\boldsymbol{\theta}} (\mathbf{z}^{(E)}_{\mathbf{x}}, T(t)) $.

Intrinsic-LoRA adapts the diffusion model to a supervised task by optimizing the LoRA adaptors on top of the self- and cross-attention layers~\cite{attention} of a single step dense predictor U-Net model. The adaptors $\boldsymbol{\theta}$ are optimized to minimize the differences between the transformed latent vector $\mathbf{z}^{(U)}_{\mathbf{x}}$ and the encoded ground-truth $\mathbf{z}^{(E)}_{\mathbf{y}} = E(\mathbf{y})$:

\vspace{-0.2cm}
\begin{equation} \label{eq:i-lora}
 \min_{\boldsymbol{\theta}} \mathbb{E}_{\mathbf{x}} [d(\mathbf{z}^{(U)}_{\mathbf{x}} , \mathbf{z}^{(E)}_{\mathbf{y}} )]  ~, \vspace{-0.1cm}
\end{equation}
where $d$ is a specific-task dissimilarity metric. Finally, the decoder $D$ transforms back  $\mathbf{z}^{(U)}_{\mathbf{x}}$ to the image space, obtaining the predicted intrinsic map $\hat{\mathbf{y}} = D(\mathbf{z}^{(U)}_{\mathbf{x}}) $. Both the encoder $E$ and decoder $D$ are frozen during training.

\vspace{-0.35cm}
\subsection{Segmentation-based Intrinsic LoRA (SI-LoRA)} \label{sec:si-lora}
\vspace{-0.15cm}

Intrinsic LoRA is built upon image-to-image generative models, thus, it handles 3-channel inputs and outputs. However, in our image segmentation problem, we need to distinguish between background and foreground, requiring a single-channel output. Hence, we define $\mathbf{y}$ as the concatenation along the third dimension of the ground-truth segmentation mask $\mathbf{m} \in \mathds{R}^{H \times W}$. Similarly, the model’s decoded output $\hat{\mathbf{y}}$ remains a 3-channel image, which we convert back into a single-channel mask $\hat{\mathbf{m}}$ by averaging across the three channels $\hat{\mathbf{y}}_c$, obtaining

\vspace{-0.6cm}
\begin{equation} \label{eq: New Obj Fn}
\min_{\boldsymbol{\theta}} \mathbb{E}_{\mathbf{x}} [\text{MSE}(\mathbf{z}^{(U)}_{\mathbf{x}} , \mathbf{z}^{(E)}_{\mathbf{y}})  ] , \quad \mathbf{y} = \mathbf{m} \otimes \mathds{1}_3 ,     \quad \hat{\mathbf{m}} = \frac{1}{3}\sum_{c=1}^3 \hat{\mathbf{y}}_c ~,
\end{equation}
where $\mathds{1}_3$ is a 3-dimensional vector of ones, and $\otimes$ denotes the outer product. Additionally, two adjustments were made to ensure effective segmentation: the dissimilarity metric $d$ between latent vectors is measured using Mean Squared Error (MSE), and the text prompt $t$ is set to "segmentation map."


\begin{figure}[t!]
  \centering
\resizebox{8.4 cm}{!} {
  \centerline{\includegraphics[width=8.5cm]{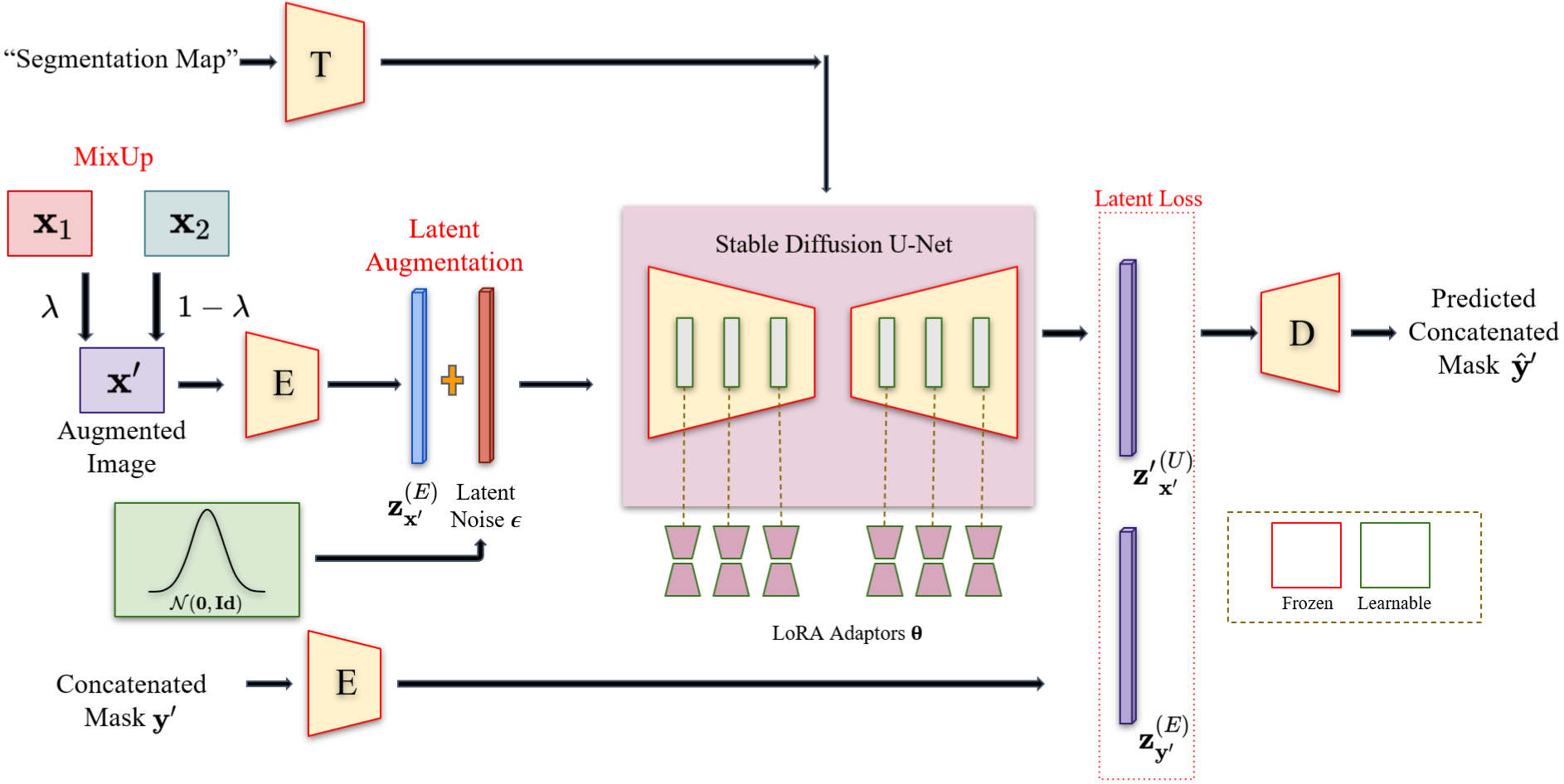}}}
  \vspace{-0.3cm}
\caption{\textbf{Segmentation-based Intrinsic LoRA (SI-LoRA) architecture with dual-space augmentation (DSA).} }
\label{fig:introd}
\vspace{-0.55cm}
\end{figure}

\vspace{-0.25cm}
\subsection{Dual-space Augmentation (DSA SI-LoRA)} \label{sec:aug}
\vspace{-0.15cm}

After successfully adapting the Intrinsic-LoRA method for our segmentation task, we shift our focus to enhance its performance. Data augmentation techniques in the image space have been widely explored and implemented to improve learning-based models~\cite{cutmix,zhang2021objectaug}. One particularly notable method is MixUp~\cite{mixup}, which generates synthetic images through the linear interpolation of multiple samples. Specifically, given two images and their corresponding labels from the training set, \((\mathbf{x}_1, \mathbf{m}_1)\) and \((\mathbf{x}_2, \mathbf{m}_2)\), MixUp produces a new augmented sample \((\mathbf{x'}, \mathbf{m'})\) as follows:

\vspace{-0.4cm}
\begin{equation}
    \mathbf{x'} = \lambda \mathbf{x}_1 + (1-\lambda) \mathbf{x}_2 ~, \quad
    \mathbf{m'} = \lambda \mathbf{m}_1 + (1-\lambda) \mathbf{m}_2 ~,
\end{equation}
where \(\lambda \in [0,1]\) is a mixing coefficient sampled from a Beta distribution with parameters $\alpha, \beta = 0.4$, which governs the interpolation between the two images and their labels.

While these techniques have proven effective, our contribution lies in extending augmentation to the latent space. Traditional diffusion models operated in the image space, however, Stable Diffusion demonstrated the effectiveness of shifting the diffusion process to the latent space~\cite{stable-diffusion}. Drawing inspiration from this and VAEs~\cite{vae}, we augment the training by parametrizing the latent vector $\mathbf{z}^{(E)}_{\mathbf{x'}}$  as a Bayesian input for the U-Net $U_{\boldsymbol{\theta}}$. In particular, the augmented $\mathbf{z'}^{(E)}_{\mathbf{x'}}$ is modeled as an isotropic Gaussian with an identity covariance matrix:

\vspace{-0.15cm}
\begin{equation} \label{eq:gaussian_noise}
  \mathbf{z'}^{(E)}_{\mathbf{x'}} \sim   \mathcal{N}\big( E (\mathbf{x'}), \mathbf{Id}  \big) ~.
\end{equation}

This probabilistic approach enhances the U-Net’s robustness by accommodating a wide range of latent inputs, instead of relying solely on deterministic encodings. To manage the stochastic nature of the sampling during backpropagation, we reparameterize the sampling process to a fixed base distribution. Consequently, the augmented latent input $\mathbf{z'}^{(E)}_{\mathbf{x'}}$ is computed by introducing a noise variable $\boldsymbol{\epsilon}$, drawn from a standard multivariate Gaussian distribution:

\vspace{-0.15cm}
\begin{equation}
\mathbf{z'}^{(E)}_{\mathbf{x'}} = E (\mathbf{x'}) + \boldsymbol{\epsilon} , \quad \boldsymbol{\epsilon} \sim \mathcal{N}(\mathbf{0},\mathbf{Id}) ~.
\end{equation}

An overview of Segmentation-based Intrinsic LoRA (SI-LoRA) with dual-space augmentation is illustrated in Fig.~\ref{fig:introd}. 


%% file: Chapters/ExperimentalResults.tex

\section{Experimental Results}

In the following section, we first present the implementation details used to successfully train our model, along with the dataset employed. Next, we provide an in-depth evaluation of the performance of SI-LoRA, including each data augmentation strategy. This is followed by qualitative assessments that highlight the effectiveness of our dual-space augmented SI-LoRA. We then compare our model with various state-of-the-art segmentation algorithms. Finally, we demonstrate its robustness by comparing its performance across different windfarms in the test set, showcasing exceptional results across diverse environments.

\vspace{-0.3cm}
\subsection{Dataset and Implementation Details}
\vspace{-0.1cm}

The dataset utilized to train (1712 images) and evaluate (320 images) the proposed method is taken from~\cite{bunet}. The input images and ground-truth segmentation masks are resized to $512 \times 512$. Decoupled regularization~\cite{adamw} with a  weight decay of $10^{-2}$, an  initial learning rate of $10^{-4}$ and a batch size of $2$ is employed. The training is stopped after $30$ epochs. For LoRA adaptors, we choose the rank $8$, consistent with the original study~\cite{intrinsic-lora}. For generating binary masks, we use Otsu's method to threshold the model predictions. The experiments were performed on an NVIDIA GeForce RTX 3090.

\vspace{-0.3cm}
\subsection{Ablation Study} \label{sec:ablation}
\vspace{-0.1cm}

Ablation studies were conducted to better understand the individual contribution of different augmentation techniques in the image and latent space. We compare four different model configurations: (1) SI-LoRA (Sec.~\ref{sec:si-lora}) without data augmentation, (2) SI-LoRA with image-space augmentation implemented in terms of MixUp~\cite{mixup}, (3) SI-LoRA with latent-space augmentation implemented in terms of noise-based probabilistic model (Sec.~\ref{sec:aug}), and (4) SI-LoRA with both image- and latent-space augmentation. Distinct metrics are evaluated to highlight the contributions of each data augmentation strategy, including the overall performance metrics of accuracy, recall, F1-score, and mean IoU (mIoU). 


Tab.~\ref{tab:ablation} showcases that applying no augmentation techniques (row 1) results in the lowest performance across all metrics, serving as a baseline for comparison. Introducing latent-space augmentation alone (row 2) shows a significant improvement in all metrics, particularly a 22.03\% increase in F1-score and a 20.48\% improvement in mIoU. This suggests that augmenting the latent space helps the model generalize better by simulating diverse, realistic variations in feature representations. When only image-space augmentation is applied (row 3), the model further boosts performance, effectively enriching the training data with more variability. Finally, combining both augmentation strategies (row 4) yields the highest performance across all metrics, with an accuracy of 99.15\%, F1-score of 98.84\%, and an mIoU of 97.69\%. These results demonstrates the synergistic benefits of applying both image- and latent-space augmentation techniques in the SI-LoRA framework. While each augmentation method contributes independently to model performance, their combined effect leads to the most significant improvements across all evaluated metrics.

\begin{table}[t!]
\centering
\caption{\textbf{Dual-space augmentation ablation study.}} 
\vspace{0cm}
\label{tab:ablation}
\resizebox{8.5 cm}{!} {
\begin{tabular}{ccccccccc}
\toprule
     \multicolumn{1}{c}{MixUp} & \multicolumn{1}{c}{Latent}  & \multicolumn{1}{c}{Accuracy} & \multicolumn{1}{c}{Recall} & \multicolumn{1}{c}{F1} & \multicolumn{1}{c}{mIoU} & \multicolumn{1}{c}{Relative} & \multicolumn{1}{c}{Relative}\\
     \multicolumn{1}{c}{\cite{mixup}} & {Noise} & {(\%)} & {(\%)}  & {(\%)}  & {(\%)} & {F1 (\%)}  & {mIoU (\%)} \\
    \midrule
    No & No & 82.30 & 73.86 & 76.48 & 76.15 & 100.00 & 100.00 \\
    No & Yes & 95.22 & 91.93 & 93.33 & 91.76 & 122.03 & 120.48 \\
    Yes & No & 97.59 & 96.75 & 97.23 & 95.24 & 127.13 & 125.05 \\
    Yes & Yes & \textbf{99.15} & \textbf{98.60} & \textbf{98.84} & \textbf{97.69} & \textbf{129.24} & \textbf{128.28} \\
\bottomrule
\end{tabular}}
\vspace{-0.3cm}
\end{table}

\vspace{-0.5cm}
\subsection{Qualitative Evaluation} 
\vspace{-0.2cm}

\begin{figure}[t!] 
\resizebox{8.5 cm}{!} {
\hspace{-0.1cm}\begin{tabular}{@{}cc@{}}
\includegraphics[width=0.5\textwidth, height=0.07\textheight]{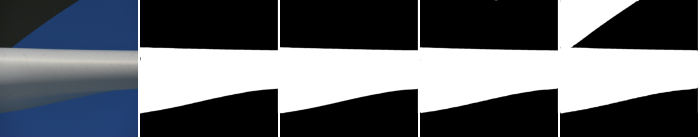}\\
\includegraphics[width=0.5\textwidth, height=0.07\textheight]{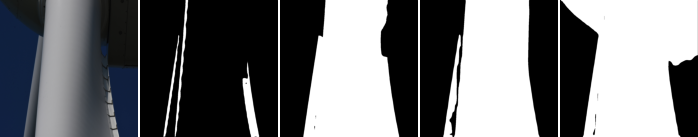}\\
\includegraphics[width=0.5\textwidth, height=0.07\textheight]{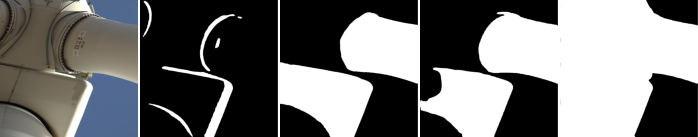}\\
\includegraphics[width=0.5\textwidth, height=0.07\textheight]{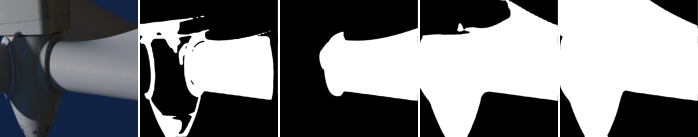}\\
\includegraphics[width=0.5\textwidth, height=0.07\textheight]{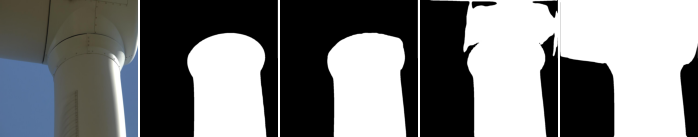}\\
\end{tabular}}
\vspace{-0.3cm}
\caption{\textbf{Qualitative comparison of distinct data augmentation strategies.} From left to right: Input image, SI-LoRA (Sec.~\ref{sec:si-lora}), SI-LoRA with image-space augmentation (MixUp~\cite{mixup}), SI-LoRA with latent-space augmentation (Sec.~\ref{sec:aug}), and SI-LoRA using both augmentations.}
\label{fig:seg-visual}
\vspace{-0.45cm}
\end{figure}

To qualitatively illustrate the segmentation masks produced by various augmentation strategies, Fig.~\ref{fig:seg-visual} presents five examples where dual-space segmentation is essential for achieving high-quality results. As discussed in Sec.~\ref{sec:ablation}, SI-LoRA struggles to generate smooth maps, leading to poor segmentation in many instances. While SI-LoRA can detect some edges of the wind turbine structure (see the second and third examples in Fig.~\ref{fig:seg-visual}), it fails to identify all the turbine components, focusing only on a single section.

When applying image- or latent-space augmentation, we observe that SI-LoRA produces smoother maps, and the generated masks include the inner regions of the wind turbine blade (WTB), rather than just outlining the structure’s edges. However, even with this improvement, SI-LoRA still falls short of fully capturing all parts of the wind turbine, as seen in all five instances. By combining both augmentation strategies, the model effectively resolves these challenging cases, generating highly accurate segmentation masks. In particular, it successfully captures all parts of the wind turbine, including darker regions on a secondary plane, as evident in the first and second examples. These results demonstrate the effectiveness of our approach in handling complex segmentation scenarios.

\vspace{-0.3cm}
\subsection{Quantitative Evaluation}
\vspace{-0.15cm}


To evaluate our proposed method, we conducted a comparative analysis as shown in Tab.~\ref{tab:sota}, benchmarking SI-LoRA against popular segmentation models. The evaluation was performed on a test set of 200 turbine images from various windfarms~\cite{bunet}. In its initial form, SI-LoRA underperformed across all metrics, lagging behind all competing models. This was primarily due to overfitting on the training masks, which significantly hindered its ability to generalize to newly acquired, unseen test images. This limitation prompted us to explore augmentation techniques aimed at enhancing its performance and generalization capabilities.

By introducing dual-space augmentation (DSA), combining MixUp~\cite{mixup} for image-space variability with noise-based probabilistic models for latent-space diversification, we developed DSA SI-LoRA. As shown in the table, DSA SI-LoRA dramatically outperforms the original SI-LoRA, surpassing state-of-the-art models by a large margin across all major metrics, except precision. These results highlight the effectiveness of our augmentation strategies in enhancing generalization and overall performance, overcoming overfitting and improving segmentation models' robustness in real-world applications. This demonstrates that pretrained generative models can be efficiently fine-tuned with limited data to perform real-world supervised tasks, such as WTB segmentation.

\begin{table}[t!]
\centering
\caption{\textbf{Quantitative comparison with competing models.} } 
\vspace{0cm}
\label{tab:sota}
\resizebox{8.5 cm}{!} {
\begin{tabular}{lccccccc}
\toprule
     \multicolumn{1}{c}{Method} & \multicolumn{1}{c}{Accuracy}  & \multicolumn{1}{c}{Precision} & \multicolumn{1}{c}{Recall} & \multicolumn{1}{c}{F1} & \multicolumn{1}{c}{mIoU} & \multicolumn{1}{c}{IoU$_{\text{bckg}}$} & \multicolumn{1}{c}{IoU$_{\text{blade}}$}\\
    \multicolumn{1}{c}{} & {(\%)} & {(\%)} & {(\%)}  & {(\%)}  & {(\%)} & {(\%)}  & {(\%)} \\
    \midrule
   SW~\cite{sw} & 93.48 & 93.57 & 91.71 & 91.37 & 87.44 & 88.64 & 86.23  \\ 
   DeepLabv3+~\cite{deeplabv3+}  & 94.14 & 96.36 & 87.38 & 89.03 & 87.47 & 90.31 & 84.62 \\  
   ResNeSt~\cite{resnest} & 94.23 & 96.84 &91.47 & 92.77 & 89.63 & 90.40 & 88.86 \\ 
   SAM~\cite{sam} & 94.36 & 97.29 & 91.22 & 92.60 & 91.66 & 92.31 & 91.01 \\ 
   DiffSeg~\cite{diffseg} & 96.37 & 83.20 & 89.74 & 85.73 & 86.40 & 91.67 & 81.13 \\ 
   U-NetFormer~\cite{unetformer} & 96.20 & 97.31 & 93.51 & 94.42 & 91.75 & 92.53 & 90.96  \\ 
   BU-Net~\cite{bunet} & 97.39 & \textbf{99.42} & 93.35 & 95.73 & 93.80 & 94.70 & 92.90  \\ 
   SI-LoRA (Sec.~\ref{sec:si-lora}) & 82.30  & 97.01 & 73.86 & 76.48  & 76.15 & 79.56 & 72.74  \\
   DSA SI-LoRA (Sec.~\ref{sec:aug}) & \textbf{99.15} & 99.20 & \textbf{98.60} & \textbf{98.84} & \textbf{97.69} & \textbf{97.56} & \textbf{97.83}  \\
\bottomrule
\end{tabular}}
\vspace{-0.5cm}
\end{table}

\vspace{-0.3cm}
\subsection{Windfarm Dissimilarity}
\vspace{-0.15cm}

\begin{figure}[t!]
  \centering
  \includegraphics[width=\linewidth]{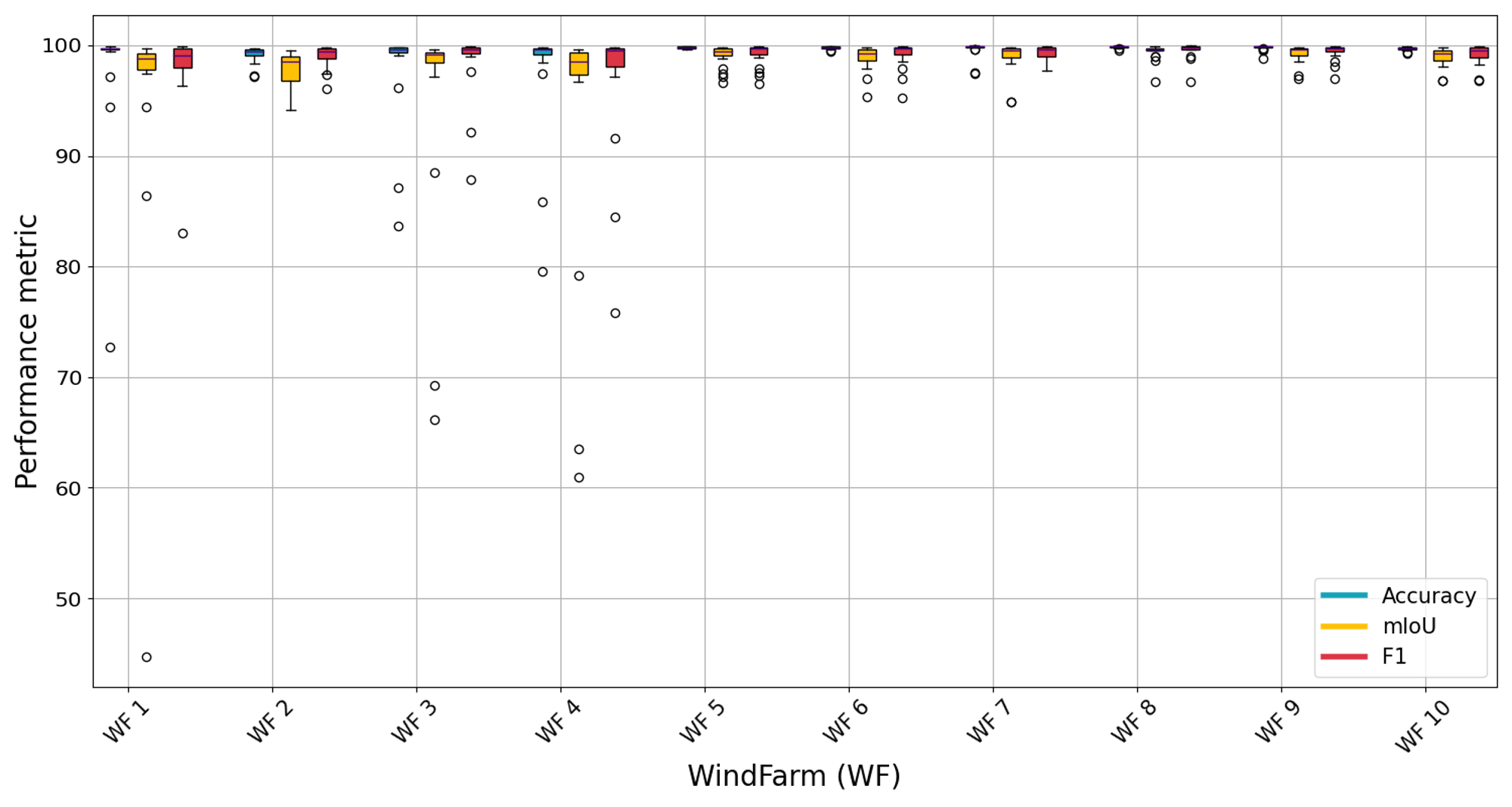}
  \vspace{-0.95cm}
\caption{\textbf{Boxplot test results across different windfarms.}}
\label{fig:boxplot}
\vspace{-0.3cm}
\end{figure}

The test dataset used in our study comprises 20 images from various windfarms~\cite{bunet}, captured using different drone configurations and locations. To evaluate the robustness of dual-space augmented SI-LoRA (DSA SI-LoRA), Fig.~\ref{fig:boxplot} presents a boxplot illustrating the performance across 10 distinct windfarms. This figure offers insights into the robustness of DSA SI-LoRA in WTB image segmentation.

The boxplot reveals consistently high average performance metrics, including accuracy, F1-score, and mIoU, with minimal variability in performance distribution. These results indicate that DSA SI-LoRA effectively generates accurate masks across a range of input environments.

\begin{figure}[t!] 
\resizebox{8.5 cm}{!} {
\hspace{-0.1cm}\begin{tabular}{@{}c|c@{}}
\includegraphics[width=0.5\textwidth, height=0.15\textheight]{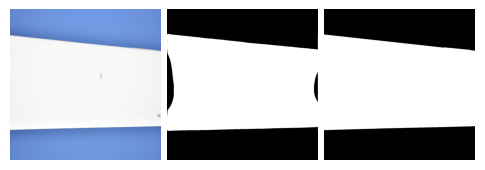}&
\includegraphics[width=0.5\textwidth, height=0.15\textheight]{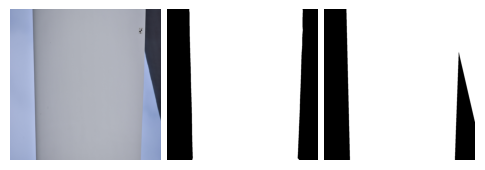}\\
\includegraphics[width=0.5\textwidth, height=0.15\textheight]{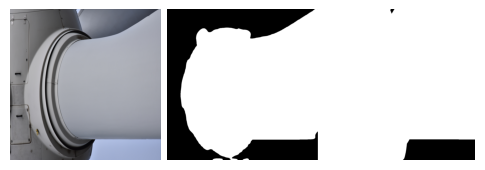}&
\includegraphics[width=0.5\textwidth, height=0.15\textheight]{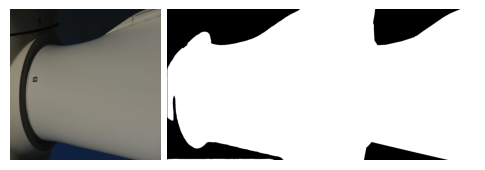}\\
\end{tabular}}
\vspace{-0.35cm}
\caption{\textbf{Failure Cases.} From left to right: Input Image,  SI-LoRA using both augmentations (Sec.~\ref{sec:aug}), and ground-truth segmentation masks. On both sides of the figure, the same information is displayed.}
\label{fig:fail}
\vspace{-0.5cm}
\end{figure}

Nevertheless, the boxplot also highlights a few outliers, with lower performance observed for windfarms 1, 3, and 4. Fig.~\ref{fig:fail} displays four representative examples where the method encounters difficulties. In these instances, high contrast between the WTB and the background, or within the WTB region itself, can lead to parts of the WTB being misclassified as background. Despite these isolated cases, they do not detract significantly from the overall robustness demonstrated by the method.


%% file: Chapters/Conclusions.tex
\vspace{-0.3cm}
\section{Conclusion}
\vspace{-0.1cm}


In conclusion, this paper presents a significant advancement in wind turbine blade (WTB) image segmentation through the development of the dual-space augmented Segmentation-based Intrinsic LoRA (SI-LoRA). By extending the capabilities of Intrinsic LoRA to image segmentation and employing an innovative dual-space augmentation strategy, our method fine-tunes generative pretrained models using minimal data, addressing the limitations of large vision universal models in specialized domains. In particular, the dual-space strategy integrates linear interpolation in the image space and probabilistic augmentation in the latent space, leading to substantial improvements in segmentation accuracy. Our experiments demonstrate that dual-space augmented SI-LoRA consistently outperforms existing state-of-the-art models in WTB segmentation, delivering robust performance across windfarms. These results highlight the potential of SI-LoRA as a powerful tool for improving the automation and reliability of wind turbine maintenance, ultimately contributing to the sustainability and efficiency of wind energy operations.

%% file: Template.bbl
\begin{thebibliography}{10}

\bibitem{delamination-efficiency}
P.~Haselbach, R.~Bitsche, and K.~Branner,
\newblock ``The effect of delaminations on local buckling in wind turbine blades,''
\newblock {\em Renewable Energy}, vol. 85, pp. 295--305, 2016.

\bibitem{turbine-failure}
Y.~Lin, L.~Tu, H.~Liu, and W.~Li,
\newblock ``Fault analysis of wind turbines in china,''
\newblock {\em RSER}, vol. 55, pp. 482--490, 2016.

\bibitem{HPL}
R.~Pérez-Gonzalo, A.~Espersen, and A.~Agudo,
\newblock ``Generalized nested latent variable models for lossy coding applied to wind turbine scenarios,''
\newblock in {\em ICIP}, 2024.

\bibitem{bunet}
R.~P\'erez-Gonzalo, A.~Espersen, and A.~Agudo,
\newblock ``Robust wind turbine blade segmentation from rgb images in the wild,''
\newblock in {\em ICIP}, 2023, pp. 1025--1029.

\bibitem{deeplab}
L.~Chen, G.~Papandreou, I.~Kokkinos, K.~Murphy, and A.~Yuille,
\newblock ``Deeplab: Semantic image segmentation with deep convolutional nets, atrous convolution, and fully connected crfs,''
\newblock {\em IEEE TPAMI}, vol. 40, no. 4, pp. 834--848, 2017.

\bibitem{sw}
X.~Pan, X.~Zhan, J.~Shi, X.~Tang, and P.~Luo,
\newblock ``Switchable whitening for deep representation learning,''
\newblock in {\em ICCV}, 2019, pp. 1863--1871.

\bibitem{deeplabv3+}
L.~Chen, Y.~Zhu, G.~Papandreou, F.~Schroff, and H.~Adam,
\newblock ``Encoder-decoder with atrous separable convolution for semantic image segmentation,''
\newblock in {\em ECCV}, 2018, pp. 801--818.

\bibitem{resnest}
H.~Zhang et~al.,
\newblock ``Resnest: Split-attention networks,''
\newblock in {\em CVPRW}, 2022, pp. 2736--2746.

\bibitem{unetformer}
L.~Wang, R.~Li, C.~Zhang, S.~Fang, C.~Duan, X.~Meng, and P.~M. Atkinson,
\newblock ``Unetformer: A unet-like transformer for efficient semantic segmentation of remote sensing urban scene imagery,''
\newblock {\em JPRS}, vol. 190, pp. 196--214, 2022.

\bibitem{attention2}
M.~Yin, Z.~Yao, Y.~Cao, X.~Li, Z.~Zhang, S.~Lin, and H.~Hu,
\newblock ``Disentangled non-local neural networks,''
\newblock in {\em ECCV}, 2020, pp. 191--207.

\bibitem{bmvc}
Y.~Meng, H.~Zhang, D.~Gao, Y.~Zhao, X.~Yang, X.~Qian, X.~Huang, and Y.~Zheng,
\newblock ``{BI-GC}onv: Boundary-aware input-dependent graph convolution for biomedical image segmentation,''
\newblock in {\em BMVC}, 2021.

\bibitem{sam}
A.~Kirillov et~al.,
\newblock ``Segment anything,''
\newblock in {\em ICCV}, 2023, pp. 4015--4026.

\bibitem{dino}
M.~Caron, H.~Touvron, I.~Misra, H.~J{\'e}gou, J.~Mairal, P.~Bojanowski, and A.~Joulin,
\newblock ``Emerging properties in self-supervised vision transformers,''
\newblock in {\em CVPR}, 2021, pp. 9650--9660.

\bibitem{intrinsic-lora}
X.~Du, N.~Kolkin, G.~Shakhnarovich, and A.~Bhattad,
\newblock ``Intrinsic lora: A generalist approach for discovering knowledge in generative models,''
\newblock in {\em CVPRW}, 2024.

\bibitem{stable-diffusion}
R.~Rombach, A.~Blattmann, D.~Lorenz, P.~Esser, and B.~Ommer,
\newblock ``High-resolution image synthesis with latent diffusion models,''
\newblock in {\em CVPR}, 2022, pp. 10684--10695.

\bibitem{mixup}
H.~Zhang, M.~Cisse, Y.~N. Dauphin, and D.~Lopez-Paz,
\newblock ``mixup: Beyond empirical risk minimization,''
\newblock in {\em ICLR}, 2018.

\bibitem{cutmix}
D.~Walawalkar, Z.~Shen, Z.~Liu, and M.~Savvides,
\newblock ``Attentive cutmix: An enhanced data augmentation approach for deep learning based image classification,''
\newblock in {\em ICASSP}, 2020, pp. 3642--3646.

\bibitem{zhang2021objectaug}
J.~Zhang, Y.~Zhang, and X.~Xu,
\newblock ``Objectaug: object-level data augmentation for semantic image segmentation,''
\newblock in {\em IJCNN}, 2021, pp. 1--8.

\bibitem{Sønderby2016a}
C.~Sønderby, J.~Caballero, L.~Theis, W.~Shi, and F.~Huszár,
\newblock ``Amortised map inference for image super-resolution,''
\newblock in {\em ICLR}, 2017.

\bibitem{jenni2019stabilizing}
S.~Jenni and P.~Favaro,
\newblock ``On stabilizing generative adversarial training with noise,''
\newblock in {\em CVPR}, 2019, pp. 12145--12153.

\bibitem{dual-space-augmentation}
F.~Zhu, Z.~Cheng, X.-y. Zhang, and C.-l. Liu,
\newblock ``Class-incremental learning via dual augmentation,''
\newblock {\em NeurIPS}, vol. 34, pp. 14306--14318, 2021.

\bibitem{lora}
E.~J. Hu, yelong shen, P.~Wallis, Z.~Allen-Zhu, Y.~Li, S.~Wang, L.~Wang, and W.~Chen,
\newblock ``Lo{RA}: Low-rank adaptation of large language models,''
\newblock in {\em ICLR}, 2022.

\bibitem{unet}
O.~Ronneberger, P.~Fischer, and T.~Brox,
\newblock ``U-net: Convolutional networks for biomedical image segmentation,''
\newblock in {\em MICCAI}, 2015, pp. 234--241.

\bibitem{clip}
A.~Radford, J.~W. Kim, C.~Hallacy, A.~Ramesh, G.~Goh, S.~Agarwal, G.~Sastry, A.~Askell, P.~Mishkin, J.~Clark, et~al.,
\newblock ``Learning transferable visual models from natural language supervision,''
\newblock in {\em ICLR}, 2021, pp. 8748--8763.

\bibitem{attention}
A.~Vaswani, N.~Shazeer, N.~Parmar, J.~Uszkoreit, L.~Jones, A.~N. Gomez, L.~u. Kaiser, and I.~Polosukhin,
\newblock ``Attention is all you need,''
\newblock in {\em NeurIPS}, 2017.

\bibitem{vae}
D.~P. Kingma and M.~Welling,
\newblock ``Auto-encoding variational bayes,''
\newblock in {\em ICLR}, 2014.

\bibitem{adamw}
I.~Loshchilov and F.~Hutter,
\newblock ``Decoupled weight decay regularization,''
\newblock in {\em ICLR}, 2019.

\bibitem{diffseg}
J.~Tian, L.~Aggarwal, A.~Colaco, Z.~Kira, and M.~Gonzalez-Franco,
\newblock ``Diffuse attend and segment: Unsupervised zero-shot segmentation using stable diffusion,''
\newblock in {\em CVPR}, 2024, pp. 3554--3563.

\end{thebibliography}
